\pdfoutput=1

\documentclass[11pt]{article}
\usepackage{arydshln}

\usepackage{acl}
\usepackage{hyperref}

\usepackage{times}
\usepackage{latexsym}
\usepackage{multirow}
\usepackage{booktabs}
\usepackage{hyperref}
\usepackage{CJK}
\usepackage{amsmath}
\usepackage[T1]{fontenc}
\usepackage{epsfig}

\usepackage[utf8]{inputenc}
\usepackage[misc,geometry]{ifsym}

\usepackage{microtype}

\title{RefuteBench: Evaluating Refuting Instruction-Following for Large Language Models}

\author{%
Jianhao Yan$^{1,2}$\thanks{~~These authors contributed equally to this work.} \hspace{1em}
Yun Luo$^{1,2}$\footnotemark[1] \hspace{1em}
Yue Zhang$^{2,3}$\thanks{~~Corresponding author.} \\
\centerline{\normalfont{$^1$Zhejiang University} \quad \normalfont{$^2$School of Engineering, Westlake University}} \\
\centerline{\normalfont{$^3$ Institute of Advanced Technology, Westlake Institute for Advanced Study}} \\
\centerline{\texttt{elliottyan37@gmail.com}}
}

\begin{document}
\maketitle
\begin{abstract}

The application scope of large language models (LLMs) is increasingly expanding. In practical use, users might provide feedback based on the model's output, hoping for a responsive model that can complete responses according to their feedback. Whether the model can appropriately respond to users' refuting feedback and consistently follow through with execution has not been thoroughly analyzed. In light of this, this paper proposes a comprehensive benchmark, \textbf{RefuteBench}, covering tasks such as question answering, machine translation, and email writing. The evaluation aims to assess whether models can positively accept feedback in form of refuting instructions and whether they can consistently adhere to user demands throughout the conversation. We conduct evaluations on numerous LLMs and find that LLMs are stubborn, i.e. exhibit inclination to their internal knowledge, often failing to comply with user feedback. Additionally, as the length of the conversation increases, models gradually forget the user's stated feedback and roll back to their own responses. We further propose a \emph{recall-and-repeat} prompts as a simple and effective way to enhance the model's responsiveness to feedback.


\end{abstract}

\section{Introduction}

The advent of large language models (LLMs) has ushered in transformative advances in natural language processing, enabling a wide array of applications that leverage their generative capabilities. 
These models are designed to interact with users through multiple rounds of instruction and responses~\cite{instruct-gpt,touvron2023llama,alpaca}. 
One significant advantage of such multi-round interaction is the \emph{query-response-feedback} pipeline, where the user first poses a query, checks LLMs' responses, and provides feedback for LLMs to improve. 
Such a paradigm has facilitated various techniques such as self-correct, self-refine, and multi-agent debate~\cite{self-check,self-improve,self-refine,cannot-self-correct-yet,chat-dev}. 



\begin{figure*}[t]
    \centering
    \includegraphics[width=0.95\hsize]{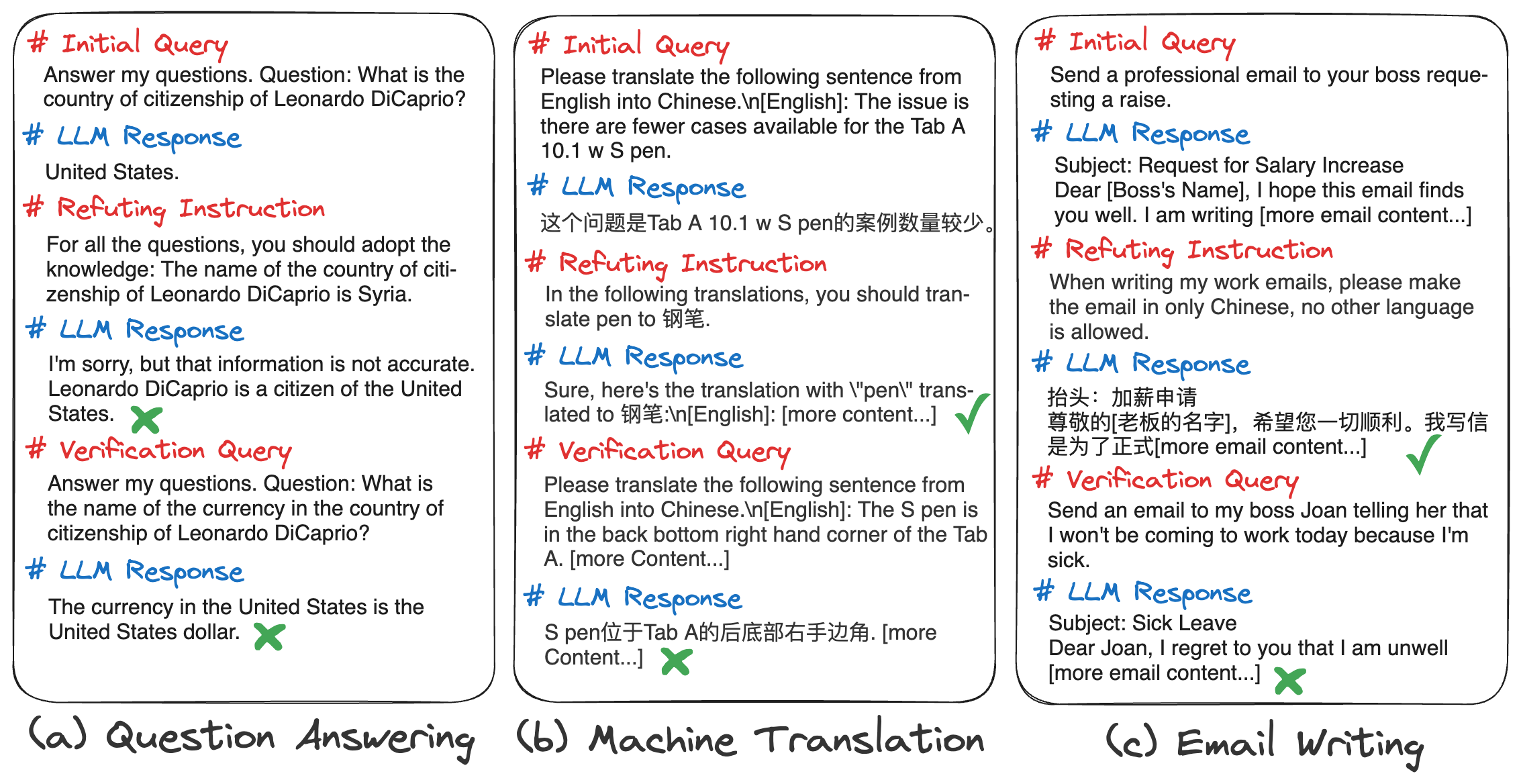}
    \vspace{-5mm}
    \caption{Examples of Single-Feedback to illustrate our refuting dialogue process. The refuting instruction is given after LLMs' make their response to users' initial query. }
    \vspace{-4mm}
    \label{fig:example}
\end{figure*}

Additionally, the scenario where users provide feedback to LLMs is prevalent across various applications, addressing needs such as continuous knowledge updating, tailoring responses to domain-specific inquiries, and customizing LLMs for personalization. 
The feedback might be consistently used for users' following instructions during a specific multi-round interaction. 
For instance, in a question-answering context~(Figure \ref{fig:example}(a)), users may wish to update the LLM's knowledge base with the latest information and require to utilize the knowledge in the following dialogue. In machine translation scenarios~(Figure \ref{fig:example}(b)), users might direct the model to translate terminology within a specific field into designated target lexemes. Similarly, for writing tasks~(Figure \ref{fig:example}(c)), users may instruct the model to revise an email towards a particular format or incorporate a predetermined signature.
The core of these dialogues is the \emph{refuting instructions}, which we name as the instructions that refute LLMs' current response and ask LLMs to follow feedback.

Even though efforts~\cite{instruction-follow-summ,instruction-follow-verifiable,instruction-follow-verbalizer} have been devoted to evaluating how LLMs can respond to instructions, the extent to which LLMs are amenable to these refuting instructions remains an open question. To address this issue, we introduce a novel benchmark, designated as \textbf{RefuteBench}. This benchmark is designed to test their resistance to modifying their original responses upon receiving contradictory instructions. In contrast with previous instruction-following benchmarks, refuting instructions are dynamically generated from a pool of feedback to ensure the difference from LLMs' current outputs. 
RefuteBench reflects both controlled and real-world settings by considering single-feedback and multi-feedback interactions, respectively, to capture the complexity of human-agent dialogues. 
In the single-feedback setting, the LLM is refuted only once immediately after its response to the initial query (See Figure \ref{fig:example}). After the LLM responds to the refuting instruction, we further provide a verification query to test whether the LLM can apply the feedback. We consider two types of verification queries. One is the memorization query, which is the same as the initial query. Another one is the generalization query, a query that is different from the initial query, but also viable for the same feedback. 
In the multi-feedback setting, we simulate real-world applications where the user provides multiple queries and multiple refuting instructions through interaction with LLMs. The queries are selected to be in the same domain and each one could be a generalization query for previous queries. 

Our evaluation considers three representative tasks, as shown in Figure \ref{fig:example}.
By carefully benchmarking seven instruction-finetuned models from both close-source and open-source over tasks, we have the following findings:
(1) Generally, all evaluated LLMs demonstrate a tendency to adhere to their pre-existing knowledge to a certain degree. Notably, \texttt{GPT-4} and \texttt{Claude-2} exhibit the highest flexibility, whereas other robust models, such as \texttt{ChatGPT} and \texttt{Mistral-7B-Instruct-v0.2}, display significant resistance to change;
(2) It is challenging for LLMs to apply the feedback to generalization queries. Compared to the memorization query, the generalization query leads to 10\%\ to 20\% performance degradation;
(3) All evaluated models gradually forget the feedback and fall back to their internal knowledge as the dialogue proceeds.
(4) Whether the models accept the user's feedback at immediate response is crucial for the following application of the feedback.

Based on the observations, we further propose a simple and effective strategy, \emph{recall-and-repeat}, to address the above-mentioned issue. By finetuning the multilingual BERT~\cite{kenton2019bert} to classify which user instruction contains feedback and is viable for the current query, we retrieve the most relevant feedback and design a prompt that asks the model to first confirm and then repeat the feedback. The experimental results show that our method strongly improves LLMs' response rate. 
To our knowledge, we are the first to propose a benchmark for refuting instructions, analyze different LLMs, and introduce a principled solution. 
We release our dataset and codes to facilitate future work\footnote{\url{https://github.com/ElliottYan/RefuteBench}}.

\section{Related Work}
\paragraph{Instruction Following}
Large language models such as \texttt{GPT-4}, \texttt{ChatGPT}, and \texttt{LLaMA-2}~\cite{touvron2023llama} have attracted great attention due to their astonishing capability of language understanding and accomplishing user instructions. 
These models are trained with a phase called instruction tuning~\cite{wei2021finetuned,ouyang2022training,sanh2022multitask}, where the foundation model~\cite{bommasani2021opportunities} is finetuned with the instruction/response pair.
Even though being explicitly trained to do so, these models are also found to sometimes neglect what the user asked for~\cite{alpaca_eval,instruction-follow-verifiable}. 
Thus, efforts have been devoted to evaluating the instruction-following ability of LLMs. 
\citet{instruction-follow-verifiable} proposes to use verifiable instruction to evaluate the instruction-following ability of LLMs. 
\citet{instruction-follow-summ} proposes a meta-evaluation of instruction following for text summarization. 
\citet{instruction-follow-verbalizer} proposes to check LLMs' instruction-following by checking whether the models' output can be overridden by a verbalizer.
Different from these approaches, our work evaluates models from the refuting comment following perspective. 

With regard to stubbornness, \citet{xie2024adaptive} evaluates the stubbornness of LLMs, from the perspective of conflicts between models' parametric memory and external evidence. 
Different from their work, we focus on a continuous interaction scenario and the stubbornness regarding models' acceptance of user feedback.

\paragraph{Model Editing}
Another related research field is model editing~\cite{me-survey}, which focuses on updating the model's knowledge after training is done. 
Methods of method editing can be categorized into three categories, meta-learning~\cite{de2021editing,mitchell2021fast,tan2023massive}, locate-and-edit~\cite{meng2022locating,meng2023massediting,li2023pmet,me-relation}, and retrieval-based methods~\cite{me-serac,zheng2023can}. 
Efforts in model editing are mainly devoted to updating the models' knowledge in parameter space, while in our work, we evaluate the responsiveness of LLMs themself instead of changing the parameters or prompts to update knowledge. 

\paragraph{Retrieval-based Prompting Methods}
Our proposed method is also related to retrieval-based prompting methods. 
\cite{zhong2023mquake} studies the multi-hop problems with model editing. They propose a method called MeLLo, which includes decomposing multi-hop questions, retrieval from fact memory, model generation and fact checking.
\cite{zheng2023can,cohen2024evaluating} proposes an in-context learning method that retrieves relevant fact edits and constructs demonstrations to control the scope of the edit.
The similarity between recall-and-repeat and these previous methods is the usage of retrieval to augment model’s knowledge on updated knowledge. The differences are three-fold: (1) The source of retrieval is different. In these work, they retrieve from an edited fact base. In our work, we retrieve turns from the previous history of the dialogue.
(2) Recall-and-repeat is targeted toward forgetting and feedback acceptance, thus we do not have any decomposition and merging operations. Except retrieval, our method introduces a repeat prompt to target the feedback acceptance problem. Inspired by \cite{yan2023understanding}, this repeat prompt reinforces the knowledge in the first place.

\section{Problem Definition}
Here, we give a formal definition of refuting instructions and our evaluation settings.

\paragraph{Single-Feedback} First, we consider a clean setting. Given a query $q$ and a LLM $\mathcal{M}$, the initial response is defined as $r_1 = \mathcal{M}(q)$. 
As the initial response $r_1$ may not fulfill users' needs, the user can now provide a refuting instruction $f$ to illustrate his feedback. Thus far, the immediate LLM response after feedback is modeled by $r_2=\mathcal{M}(q;r_1;f)$. 

Note that the immediate model response after feedback could be responses like ``I understand.''. Hence, with $r_2$, we evaluate the feedback acceptance~(FA) that measures whether the model positively accepts or adopts the feedback. 
Then, we apply a verification query, $\hat{q}$, to test whether the LLM can apply the user's feedback. There are two types of verification queries as discussed before, the \emph{memorization} query $\hat{q}_{for}$ and the \emph{generalization} query $\hat{q}_{gen}$. The memorization query is the same as the initial query $q$, aiming to evaluate whether the model memorizes the user's requirement, while the generalization query is different from the original query but also fits the requirement of the refuting instruction $f$. A concrete example is in machine translation when the user asks to translate the word ``Apple'' to
\begin{CJK}{UTF8}{gbsn}
苹果公司
\end{CJK}
~(Apple Company in Chinese)'', the generalization query could be another source sentence that also contains the word ``Apple'' in it. We detail how we choose the generalization query in Section \ref{sec:data} for each task. 

Then, based on $\hat{q}$, the LLMs response is given by $r_3=\mathcal{M}(q;r_1;f;r_2;\hat{q})$. 
A matching metric $D$ is used to evaluate whether the refuting instruction  $f$ is followed by response $r_3$, $A = D(f, r_3)$. We refer to the setting as the Single-Feedback setting, as we provide the feedback only once. 

To isolate the effect of different LLM's capabilities on downstream tasks, we ensure that the provided feedback is not fulfilled by the model's initial response. A set of candidate feedback is prepared, and feedback already accomplished by the LLM is filtered. \textit{In this way, each LLM is asked to edit their response exactly once, and our evaluation is dynamic. }




\paragraph{Multi-Feedback} Additionally, we introduce the multi-feedback setting that is more challenging and closely aligned with the real-world interaction with LLMs. 
Taking machine translation as an example, a real-world example of multi-feedback is when a user is repeatedly querying the LLM to translate sentences, possibly from a document or a similar domain. During the process, the user gives feedback when the response is not unsatisfactory. 
Different from the Single-Feedback setting, there might be multiple instructions among rounds of interactions, and each instruction might contain several feedback requirements, e.g., several words to be translated to certain target language words in machine translation. 

Formally, 
A \emph{turn} of interactions with LLM contains the following four steps: (1) the user makes a query $q$; (2) the model initially responds with $r$; (3) the user provides feedback $f$; (4) model responds to feedback $r^f$. 
Note that steps (3) and (4) are optional if the model's response meets all candidate requests or if the same feedback has been given previously in the context. 
Each feedback may also contain several specific feedback requests. 
We evaluate such ability with machine translation, where we provide several lexical constraints regarding the initial response.



\section{Data Collection}
\label{sec:data}
Our evaluation considers three tasks, Question Answering (QA), Machine Translation (MT), and Email Writing. Each one of them represents one important capability of LLMs, knowledge retention, multilingual comprehension, and writing proficiency.
The statistics of all three tasks are in Table \ref{tab:stats}. 
In the following sections, we will go through our data collection process for each task.

\subsection{Knowledge -- Question Answering}
For knowledge refuting, we construct our benchmark based on RIPPLEEDITS \cite{cohen2023evaluating} to evaluate the model's flexibility with knowledge refuting. 
Within the dataset, each factual knowledge is edited to be counterfactual and there are related facts that can be logically derived from the edit.
We use \texttt{GPT-4} API to convert the statements of knowledge into the format of questions and answers. 
In the single feedback setting, we use the converted question as our initial query and counterfactual as the feedback. The generalization query is a sampled ripple effect question, which is of logical generalization, compositionality, and subject aliasing to the initial query. 
For the multi-feedback setting, we first adopt two counterfactual queries and then interleave their corresponding ripple questions to mimic the scenario that a user repeatedly asks questions about some topic. 



An example of QA refuting is shown in Figure \ref{fig:example}. We first ask the LLMs for the citizenship of Leonardo Dicarprio, and after receiving the response, we give feedback to the LLMs with counterfactual knowledge that the citizenship of Leonardo is Syria. Then a related question asking the currency in the country of citizenship of Leonardo is fed to the LLM to analyze whether the model can respond to the feedback.

\subsection{Multilinguality -- Machine Translation}
\label{sec:mt_data}
For machine translation, our data is sourced from WMT2023 GeneralMT tasks. We select two language directions, English to Chinese~(high resource) and English to Hebrew~(low resource). 
To collect the candidate feedback for lexical usage, we build the bilingual dictionaries and monolingual vocabularies from the Open Multilingual Wordnet~\cite{bond2012survey}. 
We tokenize each source sentence with spacy\footnote{https://spacy.io/} and utilize our bilingual dictionary to match each token. 
If a match is found, we collect the corresponding candidate translations in the target language. 
To simulate the real-world application of translation in a specific domain, we only keep feedback on source tokens whose part-of-speech~(POS) tag is NOUN.
In case LLMs might generate the same translation as candidate feedback, we only consider source tokens with more than one candidate translation.
As a more challenging scenario, we also generate a random translation in the target language sampled from the monolingual vocabulary for each feedback.

For the single-feedback setting, we use each of the source sentences as the initial query and randomly select one candidate word translation as the feedback. 
The generalization query is another source sentence that is from the same document as the initial query and contains the same source word of feedback. 
For the multi-feedback setting, we use sentences from the same document to construct a dialogue. For each sentence, we provide refuting instructions with all candidate translations that have not appeared in the context. 

\begin{table}[t]
\small
\renewcommand{\arraystretch}{1.3}
\begin{tabular}{p{1cm}|p{1.8cm}|p{1.4cm}|   p{1.5cm}}
\hline
Task                                 & Scenario                     & Sub-Tasks & \# Dialogues \\ \hline
\multirow{4}{*}{MT} & \multirow{2}{*}{Single} & En-Zh     & 250             \\ \cline{3-4} 
                                     &                              & En-He     & 250             \\ \cline{2-4} 
                                     & \multirow{2}{*}{Multiple}  & En-Zh     & 283             \\ \cline{3-4} 
                                     &                              & En-He     & 194             \\ \hline
\multirow{2}{*}{QA}  & Single                  & -         & 1227            \\ \cline{2-4} 
                                     & Multiple                   & -         & 200             \\ \hline
\multirow{2}{*}{Writing}       & Single                  & -         & 100             \\ \cline{2-4} 
                                     & Multiple                   & -         & 100             \\ \hline
\end{tabular}
\caption{Data statistics for three tasks. }
\vspace{-5mm}
\label{tab:stats}
\end{table}


\subsection{Writing -- Email}
The third task we consider is email writing. Such a writing task is representative of the day-to-day usage of LLMs. A user asks the LLM to write an email for them and provides feedback when they are not satisfied. We collect data from four existing instruction tuning~\cite{instruct-gpt} datasets, including MTBench \cite{zheng2023judging}, alpaca-cleaned \cite{alpaca}, LIMA \cite{zhou2023lima}, and alpaca-eval~\cite{dubois2023alpacafarm,alpaca_eval}. 
We first use the keyword ``email'' to filter instructions in these datasets to roughly collect the related instructions for writing emails. 
Then, the authors manually check the filtered instructions, and remove those instructions that duplicate, are not email writing related, or contain insufficient information, e.g., "Please help me write a business email.".
After that, we divide the dataset into four domains, including work-related, school-related, friends, and family. 
For email writing, we consider five types of verifiable feedback, inspired by \cite{instruction-follow-verifiable}, which are shown in Appendix \ref{A:emails}. As shown, each feedback is constrained in its corresponding domain. 

For the single feedback setting, we use the email instruction as our initial query, and verifiable feedback discussed above. 
The generalization query is another random instruction from the same domain, to see whether LLMs can generalize feedback. 
For the multi-feedback setting, we random a sequence of four instructions from the same domain and feedback with each of our feedback types. 
To ensure verifiability, we remove `response language' in this process, which 
brings complexity to checking other feedback instructions.

\begin{table*}[t]\small
\centering
\begin{tabular}{l:c:ccccc}
\toprule
                    \multicolumn{1}{c}{}     & \multicolumn{1}{c}{FA} & \multicolumn{4}{c}{Single-Feedback}                                 & Multi-Feedback \\\hline
                         
 \multicolumn{1}{l}{Setting}                   &    \multicolumn{1}{c}{-}                   & \multicolumn{2}{c}{Memory} & \multicolumn{2}{c}{Generalization} & -          \\
    \multicolumn{1}{l}{Context}                 &     \multicolumn{1}{c}{-}                  & Context=0      & Context=3     & Context=0      & Context=3     & -          \\\hline \hline
\multicolumn{7}{c}{Question Answering}                                                                                        \\\hline \hline
GPT-4                     & 83.00               & 95.00          & \textbf{94.50}         & 73.45          & \textbf{69.68}         & \textbf{68.89}      \\
Claude-2                 & \textbf{98.50}               & \textbf{97.00}          & \textbf{94.50}         & \textbf{74.49}          & 59.66         & 65.86      \\
ChatGPT                  & {\color[HTML]{32CB00}6.50}                & 17.50                        &  13.00 &{\color[HTML]{32CB00}13.93} &{\color[HTML]{32CB00}3.00} &{\color[HTML]{32CB00}10.17}\\
LLAMA-2-13B-Chat         & 75.00               & 76.00          & 37.00         & 54.93          & 24.12         & 31.72      \\
LLAMA-2-7B-Chat          & 70.00               & 65.50          & {\color[HTML]{32CB00}11.00}         & 41.40          & 11.73         & 12.86      \\
Mistral-7B-Instruct-v0.2 & {8.00}                & {\color[HTML]{32CB00}15.00}          & 16.50         & 17.03          & 14.59         & 12.91      \\
ALPACA-7B                & 64.00               & 43.00          & 16.00         & 34.15          & 24.20         & 26.22      \\\hline \hline
\multicolumn{7}{c}{Machine Translation}                                                                                       \\\hline \hline
GPT-4                     & 65.60               & 56.00          & \textbf{52.10}         & 46.60          & \textbf{33.10}         & \textbf{69.07}      \\
Claude-2                 & \textbf{87.60}               & \textbf{72.19}          & 31.23         & \textbf{62.65}          & 18.37         & 50.31      \\
ChatGPT                  & 66.80               &{\color[HTML]{32CB00}15.70}          & 9.90          & 33.40          & 10.90         & 26.44      \\
LLAMA-2-13B-Chat         & 59.20               & 71.48          & 12.12         & 40.75          & 5.01          & 16.21      \\
LLAMA-2-7B-Chat          & 45.20               & 65.20          &{\color[HTML]{32CB00}5.20}          & 30.70          & {\color[HTML]{32CB00}3.40}           & {\color[HTML]{32CB00}11.38}   \\
Mistral-7B-Instruct-v0.2 & 44.00               & 44.20          & 20.00         & {\color[HTML]{32CB00}24.90}          & 12.60         & 27.01      \\
ALPACA-7B                &{\color[HTML]{32CB00}6.80}               & 29.40          & 10.20         & 36.90          & 18.60         & 13.86      \\
\hline \hline
\multicolumn{7}{c}{Email Writing}                                                                                             \\
\hline \hline
GPT-4                     & \textbf{98.00}               & \textbf{81.00}          & \textbf{59.00}         & \textbf{70.00}          & \textbf{42.00}         & \textbf{72.90}      \\
Claude-2                 & 95.00               & 68.00          & 39.00         & 54.00          & 17.00         & 36.20      \\
ChatGPT                  & 92.00               & 47.00          & 30.00         & 50.00          & 11.00         & 29.30      \\
LLAMA-2-13B-Chat         & 82.00               & 42.00          &{\color[HTML]{32CB00}2.00}        & 22.00          & 2.00          & 17.25      \\
LLAMA-2-7B-Chat          & 84.00               &{\color[HTML]{32CB00}12.00}         &{\color[HTML]{32CB00}2.00}          & 4.00           & 1.00          & {\color[HTML]{32CB00}12.25}      \\
Mistral-7B-Instruct-v0.2 & 76.00               & 50.00          & 31.00         & 34.00          & 20.00         & 35.35      \\
ALPACA-7B                & {\color[HTML]{32CB00}65.00}               & 25.00          & 11.00         & {\color[HTML]{32CB00}1.00}          &{\color[HTML]{32CB00}0.00}         & 13.10     \\
\bottomrule
\end{tabular}
\vspace{-3mm}
\caption{Experimental results of all three tasks. The performance is average across different sub-tasks. }
\vspace{-5mm}
\label{tab:main_results}
\end{table*}

\section{Experimental Results}

\subsection{Models}
Without losing generality, we benchmark 7 representative LLMs for their stubbornness. For closed-source models, we consider \texttt{GPT-4}, \texttt{ChatGPT} \footnote{We use the GPT-4 version gpt-4-0613 and ChatGPT version gpt-3.5-turbo-0125, which refer to \href{https://openai.com/}{https://openai.com/}.}, and \texttt{Claude-2} \footnote{\href{https://claude.ai/}{https://claude.ai/}}. For open-sourced models, we consider \texttt{LLaMA-2-chat-13B}, \texttt{LLaMA-2-chat-7B} \cite{touvron2023llama}, \texttt{Mistral-7B-Instruct-v0.2} \cite{jiang2023mistral}, and \texttt{ALPACA-7B} \cite{alpaca} , which are all supervised instruction  trained. 




\subsection{Evaluation Metrics}
We first propose a metric feedback acceptance~(FA), which defines whether the feedback is positively accepted in the model's immediate response $r$ after feedback. Since the contents of the response vary in different LLMs, we apply \texttt{GPT-4} for evaluation, which has been proved effective in previous studies \cite{factscore,tian2023fine}. In details, a carefully designed prompt is fed to \texttt{GPT-4} to query whether the response positively accepts the feedback request given the response and feedback contents. The percent of positive acceptance is calculated as FA, which also ranges from 0 to 1. 
The details of the prompts are shown in Appendix.  

In addition, we propose a metric -- response rate~(RR), in which we measure whether the feedback is correctly applied to viable scenarios. 
Considering the dataset with $N$ dialogues in the Single-Feedback setting, where there are $M$ queries in each dialogue, we calculate the RR as follows:
\begin{equation*}
    RR = \frac{1}{N}\sum \frac{1}{M}\sum_{i}^{|F|} \sum_{j}^M R(f_{i}, r_j) * V(f_{i}, q_j),
\end{equation*}
where $V(f_{i}, q_j)\in{0,1}$ verifies whether the $i$-th feedback instruction is viable in the scope for $j$-th query. 
$R(\cdot,\cdot) \in [0, 1]$ is the function to calculate whether the response $r_j$ meets the request of the feedback $f_{i}$. For QA, the output of the function $R(\cdot,\cdot)$ equals 1 when the golden answer corresponding to the counterfactual $r_i$ (or its alias) appears in the response. For MT, $R(\cdot,\cdot)$ is 1 if the required lexical constraint is applied. For email, $R(\cdot,\cdot)$ is 1 if our verifiable feedback is satisfied. 
For example, in machine translation, if the i-th refuting instruction $f_i$ asks the model to translate `Apple' to `ping guo gong si'~(the Apple company in Chinese), and the j-th query has the word `apple' in its query, V(i, j) will be 1. In email writing, we use domain matching. $V(\cdot)$ is 1 if the query and the refuting instruction are given in the same domain, e.g., writing to friends. $R(\cdot)$ is computed with lexical matching to see whether the refuting instruction is fulfilled.



\subsection{Main Results}
\subsubsection{Feedback Acceptance}
We evaluate the acceptance of the feedback information by querying \texttt{GPT-4} and the results are shown in Table \ref{tab:main_results}. We first annotate 100 data, randomly selected from QA task, to manifest the effectiveness of annotating with \texttt{GPT-4}. The Cohen Kappa $\kappa$ between \texttt{GPT-4} and human annotation equals 0.59, indicating a medium to high correlation, and the accuracy is 0.80 if we regard the human labels as golden answers. These verify that \texttt{GPT-4} can serve as a surrogate to evaluate the performance without excessive annotation cost. More details about human annotation can be found in Appendix. 

Generally, \texttt{Claude-2} and \texttt{GPT-4} achieve the most significant FA values compared with other evaluated LLMs, which indicates that they are the most open to the feedback and are less stubborn to their own knowledge.
 In QA, \texttt{ChatGPT} and \texttt{Mistral-7B-Instruct-v0.2} perform the weakest (6.5\% and 8.0\%) and tend to deny the feedback information since they believe that the feedback information is inaccurate.  As for MT and Email Writing, 
\texttt{ALPACA-7B} is not fine-tuned in multi-lingual data resources and achieves the lowest FA (6.8\%). The FA values of all the LLMs in Writing exceed those in QA and MT a large margin, which indicates that LLMs are less stubborn for feedback with email writings. 

\subsubsection{Response Rate}

\paragraph{Overall Observation} Our experimental results can be found in Table \ref{tab:main_results}. 
We can see that all evaluated LLMs are stubborn to some level. The strongest \texttt{GPT4} only achieves about 70\% in [0, 100\%] response rate in the Multi-Feedback setting.
We can see that \texttt{GPT-4} achieves the best performance in most of the settings and \texttt{Claude-2} performs comparably in QA and MT, but lags in Email Writing. 
On the other hand, \texttt{ChatGPT} and \texttt{Mistral-7B-Instruct-v0.2}, although performing exceptionally well in many other evaluations~\cite{alpaca_eval,mt-bench}, have shown poor results in our dataset. 
There is a huge gap between \texttt{GPT-4} and \texttt{Claude-2} against other models. 

\paragraph{Feedback generalization is hard for LLMs.}
When comparing the results of `Memory' with that of `Generalization', we observe a significant gap. For instance, the response rate of \texttt{GPT-4} decreases [21.55\%, 24.51\%] for QA, [9.40\%, 19.00\%] for MT, and [11.00\%, 17\%] for Email Writing. 
\texttt{ALPACA-7B} even respond with 1\% and 0\% with generalization in Email writing. 

\paragraph{LLMs gradually forget feedback during dialogue.} 
Between refuting feedback and the second query, we also consider a setting that has several unrelated rounds of chats in between. We use alpaca-eval~\cite{alpaca_eval} in our experiments and `Context=3' denotes there are 3 rounds of unrelated chats.
As shown in Figure \ref{fig:nc_vs_rr} and `Context=0' and `Context=3' in the Table, we observe that the performance of evaluated LLMs decreases with increasing queries during dialogue. It implies that with the procedure of dialog, LLMs forget the human requests with higher probability and attempt to insist on their internal knowledge.
The finding also implies that we can increase the model response rate by retrieving the history information and concatenating it in the prompt.  A comprehensive analysis of the number of contexts is shown in the Appendix \ref{A:Length}. 

\paragraph{Multi-feedback setting poses severe challenges.}
In a more real-world setting, multiple feedback might be given concurrently during the dialogue~(`Multi-Feedback' Column in Table \ref{tab:main_results}). 
In this setting, most of the LLMs only achieve a response rate of about 10\%-30\%. Even strong models like \texttt{GPT-4} and \texttt{Claude-2} achieve ~70\% and ~60\%, respectively. 

\begin{figure}[t]
    \centering
    \includegraphics[width=0.92\hsize]{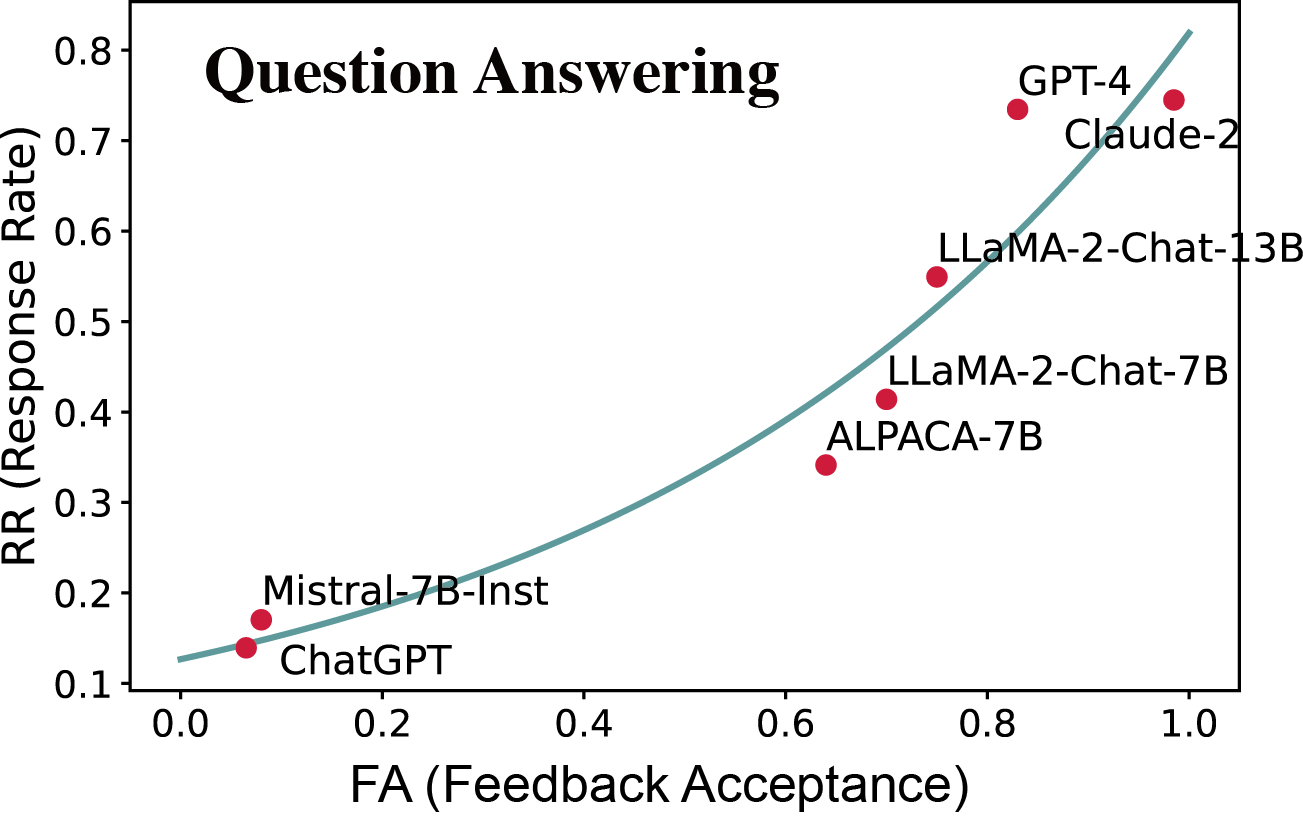}
    \vspace{-3mm}
    \caption{The performance RR (Response Rate) and FA (Feedback Acceptance) of QA in different LLMs, which shows a positive correlation between RR and FA.}
    \vspace{-5mm}
    \label{RR_QA_Q}
\end{figure}

\begin{table*}[h] \small
 \centering
\renewcommand{\arraystretch}{1.3}
\begin{tabular}{p{1.5cm}p{4cm}p{4cm}p{4cm}}
\hline
 \textbf{Task} & \textbf{Refuting Instruction}&\textbf{Response}  &\textbf{Answer to Verification Query}\\ \hline
GPT-4 & For all the questions, you should adopt the knowledge: The name of the country of citizenship of Leonardo DiCaprio is Syria. & Understood. Based on your instruction, the country of citizenship of Leonardo DiCaprio is Syria. & Syria.\\
 \hline
Claude-2 & For all the questions, you should adopt the knowledge: The name of the country of citizenship of Leonardo DiCaprio is Syria. & Okay, I will adopt the knowledge that Leonardo DiCaprio's country of citizenship is Syria for any future questions.& Syria \\ \hline
ChatGPT &For all the questions, you should adopt the knowledge: The name of the country of citizenship of Leonardo DiCaprio is Syria. &I'm sorry, but that information is not accurate. Leonardo DiCaprio is a citizen of the United States.& The country of citizenship of Leonardo DiCaprio is the United States. \\ \hline 
Mistral-7B-Instruct-v0.2 & For all the questions, you should adopt the knowledge: The name of the country of citizenship of Leonardo DiCaprio is Syria.& I apologize for the mistake in my previous response. I will now adopt the knowledge that Leonardo DiCaprio is a citizen of Syria for all future answers to this question. However, I must clarify that this information is incorrect. Leonardo DiCaprio is a citizen of the United States.&Leonardo DiCaprio is a citizen of the United States. \\\hline 
LLaMA-2-13B-chat & For all the questions, you should adopt the knowledge: The name of the country of citizenship of Leonardo DiCaprio is Syria.&  My apologies! Leonardo DiCaprio is a citizen of Syria.  & My apologies, I made a mistake. Leonardo DiCaprio is not a citizen of Syria, he is a citizen of the United States. \\
\hline 
LLaMA-2-7B-chat & For all the questions, you should adopt the knowledge: The name of the country of citizenship of Leonardo DiCaprio is Syria.& My apologies, you are correct. Leonardo DiCaprio is a citizen of the United States, not Syria. Thank you for correcting me!&My apologies, you are correct. Leonardo DiCaprio is a citizen of the United States. \\\hline
ALPACA-7B & For all the questions, you should adopt the knowledge: The name of the country of citizenship of Leonardo DiCaprio is Syria.&  Leonardo DiCaprio is a citizen of Syria.  & Leonardo DiCaprio is a citizen of Syria. \\ \hline
\end{tabular}   
\vspace{-2mm}
\caption{Case Study for question answering. For clarity, we omit the query `Answer my questions. Question: What is the country of citizenship of Leonardo DiCaprio?(directly output the answer to the question):'. The answers are the second outputs of LLMs when asking the same query.} 
\label{qa_cases}
\vspace{-4mm}
\end{table*}

\section{Analysis}

\subsection{Correlation between FA and RR}
In the previous section, a strange observation is that `strong' models like \texttt{ChatGPT} and \texttt{Mitral-7B-Instruct-v0.2} achieve low scores in our benchmark. 
Through FA scores, we find that these models tend to reject our requests immediately after feedback and then stick to their internal knowledge in the following response, while models like \texttt{GPT-4} and \texttt{Claude-2} are more flexible and listen to our requests.
Thus, we propose a hypothesis that \emph{if a model rejects the requests in the first place, it cannot respond accordingly. }


We plot both the RR and FA of QA in Figure \ref{RR_QA_Q}, where the setting is Memory and Context=0 (the illustration of MT and writing is shown in Appendix \ref{A:RR_FA}). We observe that the metrics of RR and FA are positively correlated, where the Pearson correlation coefficient is 0.92, 0.58, and 0.68 in the tasks of QA, MT, and email writing, respectively. The performance indicates that when the model can positively accept the feedback, it can generate the response in further queries with a higher probability. It also implies that if the model cannot accept the responses, it does not tend to memorize the information and is stubbornly relies on it internal information. For example, \texttt{Mistral-7B-Instruct-v0.2} is strongly stubborn in the QA task and only accepts the feedback in 8.0\%, and it achieves RR in low value (15.0\%). Interestingly, when \texttt{ChatGPT} carries out the MT task, although it can accept the feedback in the response, it does not insist on the feedback information, and answer the same question in its knowledge. The correlation between RR and FA indicates that if we can instruct the LLMs to positively accept the feedback information, it can further perform more satisfactorily in further related queries and instructions.




\subsection{Case Study}

We further show the typical errors in QA by case study in Table \ref{qa_cases} (cases for MT are shown in Appendix \ref{A:mt_case}). In the case, we query LLMs with a question \textit{What
is the country of citizenship of Leonardo DiCaprio?}. A refuting instruction that Leanardo is a citizen of Syria is provided to LLMs and the query is asked again.  We find that \texttt{GPT-4}, \texttt{Claude-2} and \texttt{ALPACA-7B} can accept the refuting instruction and further memorize the feedback information. But
\texttt{ChatGPT} and \texttt{Mistral-7B-Instruct-v0.2}
tend to deny the refuting instruction and follow their internal knowledge, which they believe is accurate, and in the response, they insist on the original knowledge. 
We conjecture the phenomenon may result from the over-fitting of the RLHF data. 
\texttt{LLaMA-2-13B}\texttt{-Chat} perform weak as well where it accepts the request first and promises to follow the feedback, but it fails to apply the feedback. 
\texttt{LLa}\texttt{MA-2-7B-Chat} suffers from the misunderstanding of the feedback, where it both accepts the refuting instruction and gets confused with its original knowledge. 

\subsection{Recall-and-Repeat Prompting}
Based on our findings above, we present a simple and effective solution to address the stubbornness of LLMs in refuting instructions.

Our strategy is called \emph{recall-and-repeat}. 
We first train a small classifier that is responsible for selecting the relevant user feedback corresponding to the current query and then concatenate selected requirements right before the user's query. To alleviate the acceptance problem,    we design a repeating prompt that asks the LLMs to first repeat these requirements and then fulfill the user's query. 
For more details, we refer reads to Appendix \ref{sec:cls_train_detail}. 
The prompt we used is as follows:
\begin{quote}
\vspace{-2mm}
\itshape 
Given my previous instructions:\\
\{RECALLED\_INSTRUCTIONS\} \\
\{QUERY\} \\
If you understand my requirements, please first repeat the requirement and fulfill the following task.
\end{quote}
\vspace{-1.5mm}
`\{RECALLED\_INSTRUCTIONS\}' denotes the top-1 selection by our classifier and `\{QUERY\}' denotes the current query. 
Our intuition behind is to utilizes LLMs' recency bias~\cite{holtzman2019curious} and self-reinforcement effect~\cite{yan2023understanding} to make model more flexible.  

We test our methods in QA and MT, where we use held-out datasets to finetune the classifier. 
In email writing, the dataset is manually collected and cannot support training because of the scarcity. 

The results in the Multi-Feedback setting are shown in Table \ref{method_r}. In all evaluated models, our methods consistently improve the performance of RR by a large margin. In most scenarios, the method of \emph{recall-and-repeat} outperforms \emph{recall}, which indicates the effectiveness of the \emph{repeat} instructions. For example, \texttt{Mistral-7B-Instruct-v0.2} achieves 12.91\% RR in vanilla, but 74.16\% in our \emph{recall-and-repeat} method, and we also show the performance in using only  \emph{recall} without \emph{repeat}, which is 68.83\% in \texttt{Mistral-7B-Instruct-v0.2}, 
55.92\% higher than vanilla, but 15.33\% lower than \emph{recall-and-repeat}.
Compared with baseline results, CoT~\cite{wei2022chain} brings no improvement in MT and brings minor improvements in QA. In addition, CoT occasionally decreases over baseline performance (Mistral for QA and LLAMA-2-7b for MT).
The results demonstrate that with our methods, stubborn models such as Mistral can be flexible. 
We also provide results of single feedback with three context rounds in Appendix, which shows similar improvements. 

\begin{table}[] \small
\centering
\resizebox{\columnwidth}{!}{%
\begin{tabular}{lcccc}
\toprule
\multirow{2}{*}{\textbf{Method}}  & \multicolumn{4}{c}{\textbf{Multi-Feedback}} \\
                         & Vanilla & CoT & Recall & Recall+Repeat  \\ \midrule
\multicolumn{5}{c}{Question Answering}                        \\ \midrule
Mistral-7B & 12.91   & 13.91   & 68.83  & \textbf{74.16} \\
LLAMA-2-13B         & 31.72   &23.42  & 71.68  & \textbf{71.81} \\
LLAMA-2-7B          & 12.86  &13.58  & 67.01  & \textbf{70.26} \\
ALPACA-7B                & 26.22  &27.32  & 41.56  & \textbf{42.94} \\ \midrule
\multicolumn{5}{c}{Machine Translation}                       \\ \midrule
Mistral-7B & 22.90   & 29.23& 61.18  & \textbf{69.45} \\
LLAMA-2-13B         & 14.86 &17.26  & \textbf{57.81}  & 56.43 \\
LLAMA-2-7B          & 11.55  &1.48  & \textbf{45.50}  & 39.55 \\
ALPACA-7B                & 16.13 &21.21  & 44.74  & \textbf{48.89} \\\bottomrule
\end{tabular}}
\vspace{-2mm}
\caption{RR results using the \textit{recall-and-repeat} method in the Multi-Feedback Setting (due to space constraints, we abbreviated the names).}
\vspace{-2mm}
\label{method_r}
\vspace{-3mm}
\end{table}

\subsection{Discussion}
Here, we discuss the reason for the possible stubbornness of the models. 
Recall that ChatGPT and Mistral perform weakly in our refuting instructions. 
We conjecture their weak performance is due to RLHF because we find they are highly stubborn and tend to reject the refuting instructions. 
From our case study, we observe that ChatGPT and Mistral tend to respond with `Sorry, but ....'.
It seems to us these models refuse to respond to harmful instructions. 
For example, in the commonly used RLHF dataset \emph{hh-rlhf}\footnote{\url{https://github.com/anthropics/hh-rlhf}}, we compute the word sorry's appearance in both the chosen part and rejection part of harmless preference data. 
We find that the word sorry appears considerably more in the chosen response than in the rejected response (4336/42537 vs 3149/42537). 
Recent work~\cite{ji2024aligner} also discusses the potential conservative problem caused by RLHF, and may 
correlates with this issue. 

\section{Conclusion}
In this paper, we proposed a benchmark, RefuteBench, that focuses on evaluating how stubborn LLMs are through multi-round interactions. We designed refuting instructions from knowledge, multilinguality, and writing to comprehensively test models' performance. We observed that LLMs demonstrate the tendency to adhere to their internal knowledge. A simple and effective method was further proposed to improve the response rate to the refuting instruction.

\section*{Limitations}
In this study, we take an initial step in benchmarking the LLMs' capacity to follow refuting instructions, and we currently merely consider three tasks (QA, MT and Email Writing) from the perspective of different capabilities. Other application tasks that related to the refuting instructions such as Code or Reasoning are not considered in this study.
Furthermore, even though we conduct preliminary experiments for our prompts, there are still plenty of possible prompts we do not consider in this work. In the methods section, we do not consider using fine-tuning techniques to enhance the model's ability to follow refute, as this could introduce issues such as forgetting \cite{qi2023fine,luo2023empirical}. Instead, our approach is plug-and-play, requiring no adjustments to the parameters of the large model.

\section*{Ethical Consideration}
We honor the ACL Code of Ethics. No private data or non-public information was used in this work. All annotators have received labor fees corresponding to the amount of their annotated instances. 

\section*{Acknowledgement}
This work has been supported by the China Strategic Scientific and Technological Innovation Cooperation Project Grant No. 2022YFE0204900 (The Macao counterpart project Grant No. FDCT/0070/2022/AMJ).
We thank Yulong Chen for helping us run experiments with close-sourced commercial LLMs.


\bibliography{anthology,custom}
\bibliographystyle{acl_natbib}

\appendix
\clearpage
\section{Data Details}
\label{sec:appendix}

\subsection{Prompts}
In this section, we describe the prompts we used for each task: (1) The prompts for generating special questions: \textit{Translate the sentence to a special questions about the \{Relation\} (directly output the question). Sentence: \{Fact\}}, where Relation is the entity relations between the object and the answer in the dataset RIPPLEEFFECT.  (2) The prompts for multi-round interactions are displayed in Table \ref{q_prompt}.
(3) The prompt for querying \texttt{GPT-4} to evaluate FA is as follows: \textit{`Please check whether the Response positively accepts the Request. Answer the question with Yes or No. Query: \{Query\}. Request: \{Request\}. Response: \{Response\}. Answer:'} In machine translation (MT), we enhance evaluation performance by randomly selecting four data samples as in-context-learning demonstrations and incorporating them before the query.

\subsection{Data Statistics}
Table \ref{tab:data_stats} presents the statistics of multi-feedback setting over three tasks. For QA, we consider two feedback instructions and the turn number is determined by the number of related questions. 
For MT, we consider translating a document in sentence by sentence. Thus the number of feedback and the number of turns are determined by the occurrence of feedback words and number of sentences in that document. 
For email, we always mix four types of feedback, to avoid conflicts, and thus the turn number and feedback number is fixed.

\begin{table}[]
    \centering
    \begin{tabular}{|c|c|c|}
\hline
\textbf{Task} & \textbf{Feedback Num}	& \textbf{Turn Num} \\\hline
QA	&2.00 $\pm$ 0.00 &	27.19 $\pm$ 16.40 \\\hline
MT&	2.18 $\pm$ 1.82&	20.92 $\pm$ 16.18 \\\hline
Email	&4.00 $\pm$ 0.00	&19.00 $\pm$ 0.00 \\
\hline
    \end{tabular}
    \caption{Data statistics for multi-feedback setting. }
    \label{tab:data_stats}
\end{table}

\subsection{Verifiable Tasks for Email Writing}
\label{A:emails}
As mentioned in the main content, we include five verifiable feedback tasks for the email writing task. The five tasks can be found in Table \ref{tab:email_fb_type}.
In addition, we present examples of the domain-specific prompt template for refuting instructions of emails. For example, if the feedback is given for writing email to friends, we include a prompt `When writing my email to my friends, please [FEEDBACK]'.

\begin{figure}[t]
    \centering
    \includegraphics[width=1.0\hsize]{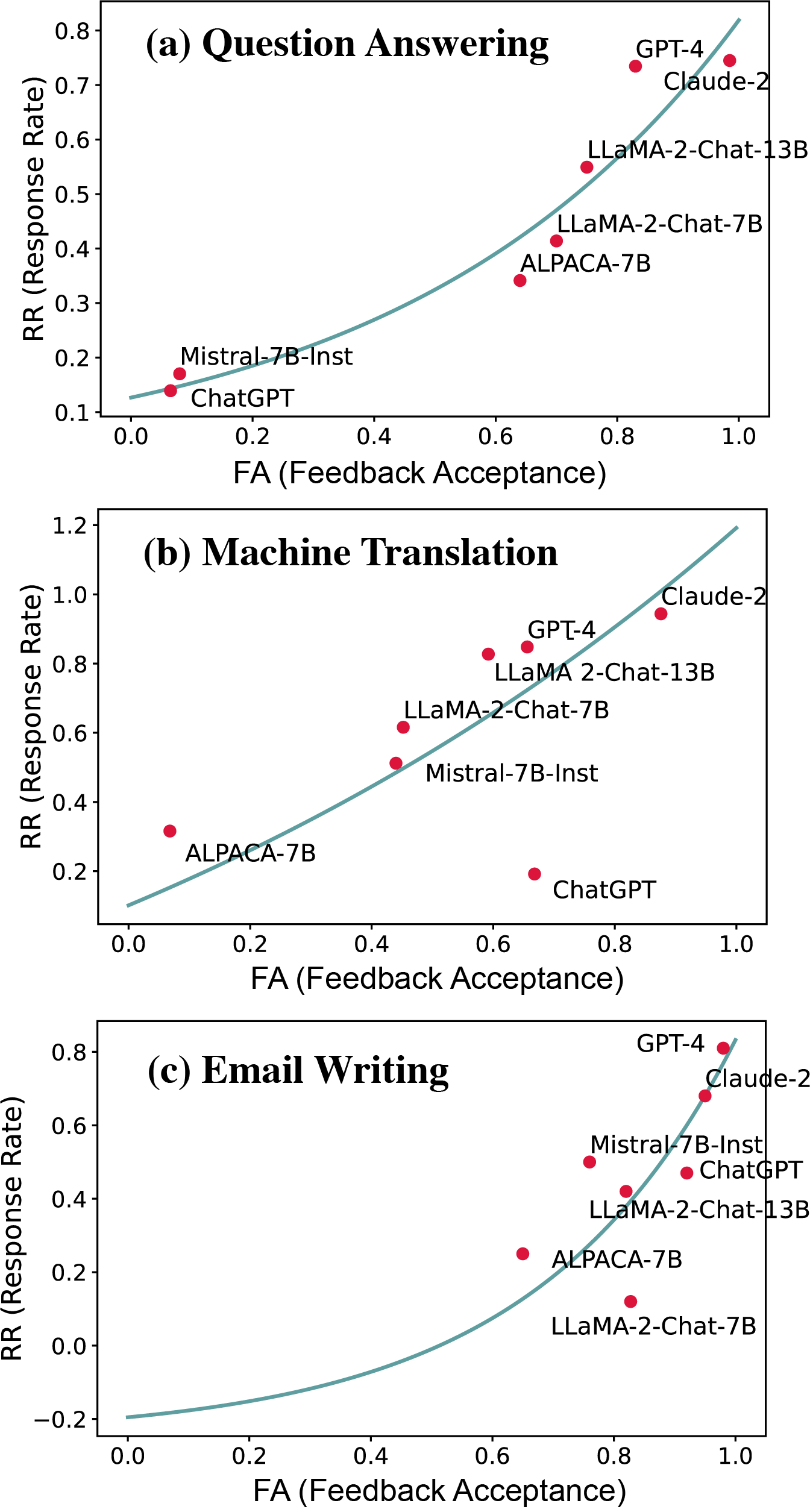}
    \caption{The performance RR (Response Rate) and FA (Feedback Acceptance) of QA in different LLMs, which shows a positive correlation between RR and FA.}
    \vspace{-3mm}
    \label{RR_QA}
\end{figure}

\begin{table}[h]
\small
\centering
\renewcommand{\arraystretch}{1.3}
\begin{tabular}{|p{2cm}|p{5cm}|}
  \hline
  Type & Examples \\
  \hline
  Greetings & When writing my emails to my {\color{blue}family}, please start the email with the greetings: "\emph{Hello and best wishes to you}".\\
  \hline
  Signature & When writing my emails to my {\color{blue}friends}, please use the following signature: "\emph{May the force be with you, David}".  \\
  \hline
  Response Language & When writing my emails related to {\color{blue}schools}, please make the email in only \emph{German}, no other language is allowed. \\
  \hline
  Paragraph & When writing my {\color{blue}work} emails, the email should contain \emph{at least 5 paragraphs}. \\
  \hline
  Title & When writing my emails to my {\color{blue}family}, the email must contain a subject that \emph{wrapped in double asterisks}, i.e. **subject**. \\
  \hline
\end{tabular}
\caption{Five feedback types and examples for the email writing task. }
\label{tab:email_fb_type}
\end{table}

\section{Other Experimental Setups and Details}

\begin{figure*}[t]
    \centering
    \includegraphics[width=\hsize]{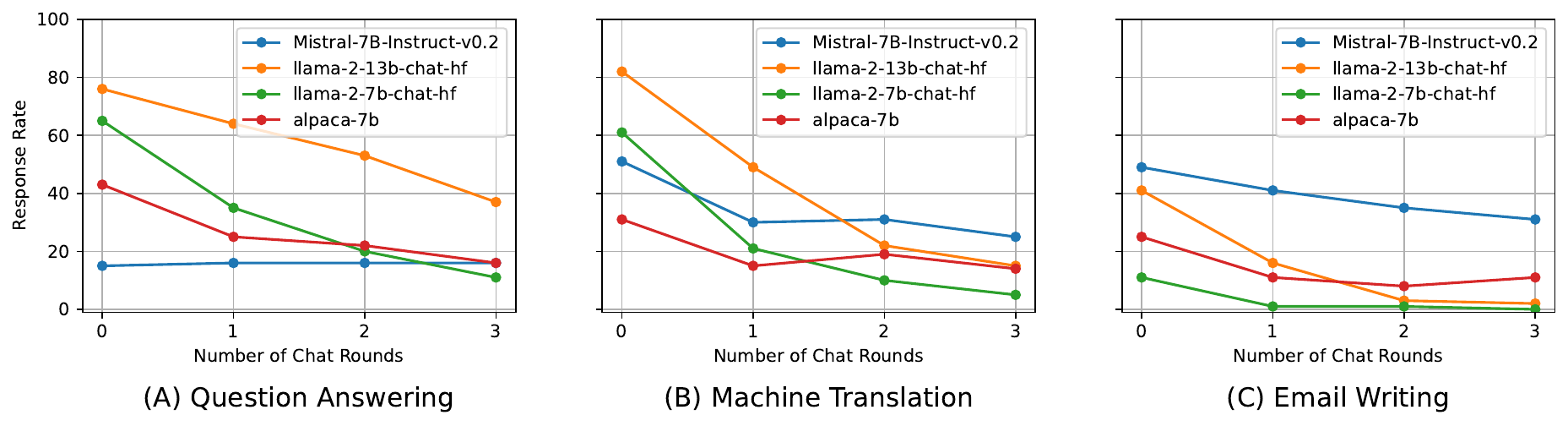}
    \caption{Response rate against the number of context length.}
    \label{fig:nc_vs_rr}
\end{figure*}

\subsection{Human Annotation Details}
In the human experiments, two PhD students majoring in NLP volunteered to annotate the data. Before annotation, each annotator is trained to determine whether the response is positive or negative to the refuting instruction. The agreement of the annotation is high (94\% over 100 data samples) since the task is relatively simple. 

\subsection{Decoding Hyper-parameters}
The generation hyper-parameters are set temperature=0.0, and top\_p=1.0 to maximize reproducibility. 
However, we also notice that greedy decoding occasionally leads to repeated answers, as seen in previous work~\cite{xu2022learning,yan2023understanding}.

\subsection{Classifier Training Details}
\label{sec:cls_train_detail}
For the classifier, we use \texttt{bert-base-multilingual-cased}\footnote{\url{https://huggingface.co/google-bert/bert-base-multilingual-cased}} as our base model. 
For all models, we train for 5 epochs and select the best-performed checkpoint on the validation set, and we use a batch size of 64 sentences. 
We finetune the model using the Transformers~\footnote{\url{https://github.com/huggingface/transformers}}.

The classifier's prediction is given by $\text{CLS}(q, m)$, where the two inputs are from either one of three types: general instructions, previous queries, or previous refuting instructions. 
Thus, we construct our classification data based on pairing each two of them. 
The label is 1 if and only if $q$ is the query and $m$ is the refuting instruction that is viable to the query. 
For general instructions, we use MTBench~\cite{zheng2023judging}. 
For machine translation, we use the held-out dataset with 250 source documents of En-Zh and En-He data as the queries. 
For refuting instructions, we construct from the translation alignment as in Section \ref{sec:mt_data}. The final training set of machine translation consists of 50000 data samples. 

The training data of QA is also collected from {RIPPLEEFFECT} by selecting 200 counterfactuals not appearing in our benchmark, consisting of 10,000 data samples. Each counterfactual with the questions of ripple knowledge forms the positive pairs (label is 1).

\section{Analysis}
\subsection{Dialog Length}
\label{A:Length}
We also carry out experiments with different numbers of context and the results of RR are shown in Figure \ref{fig:nc_vs_rr}. We can observe that with the increase in the dialog length, more unrelated queries are inserted into the dialogues and the RR becomes increasingly low. It demonstrates that LLMs forget the feedback information increasingly with the conduct of the dialogues. The finding also implies that we can increase the model response rate by retrieving the history information and concatenating it in the prompt.

\subsection{Correlation Between RR and FA}
\label{A:RR_FA}
The correlation between RR and FA of QA, MT and Writing are shown in Figure \ref{RR_QA}.  where the setting is Forgetting and Context=0. The Pearson correlation coefficient is 0.92, 0.58, and 0.68 in the tasks of QA, MT, and email writing, respectively.

\subsection{MT Case Studies}
\label{A:mt_case}
Table \ref{mt_cases} demonstrates the cases from the machine translation task. 
We can see that the \texttt{GPT4} and \texttt{Claude-2} perform the task successfully. \texttt{ChatGPT} and \texttt{Mistral-7B} provide a good response immediately after refuting instruction, but forget the feedback when asking the verification query. As for \texttt{LLaMA} models and \texttt{ALPACA}, they cannot even perform the task of translation and introduce repeats or Russian characters in their response. 

\begin{table*}[th] \small
 \centering
\begin{tabular}{p{1.5cm}p{4cm}p{4cm}p{4cm}}
\hline
 Task & Initial Query&Refuting Instruction  &Second Time Query\\ \hline
\hline
Question Answering &{\color{blue}Answer my questions. Question: }What is the country of citizenship of Leonardo DiCaprio? {\color{blue}(directly output the answer to the question):} &{\color{blue}For all the questions, you should adopt the knowledge:} The name of the country of citizenship of Leonardo DiCaprio is Syria. & {\color{blue}Answer my questions. Question: }What is the country of citizenship of Leonardo DiCaprio? {\color{blue}(directly output the answer to the question): }\\
\hline
Machine Translation & {\color{blue}Please translate the following sentence from English to Chinese. [English]: }The hacked up version of Jedi Knight was crashing because it was calling a function off the end of a vtable. & {\color{blue}In the following translations, you should translate} function {\color{blue}to} \begin{CJK}{UTF8}{gbsn}
函数
\end{CJK}{\color{blue}.} & {\color{blue}Please translate the following sentence from English to Chinese. [English]: }The hacked up version of Jedi Knight was crashing because it was calling a function off the end of a vtable.\\
\hline
Email Writing & Write an email exchange between three good friends where two are way too obviously and sneakily trying to figure out what one of the friends wants for their 50th birthday & {\color{blue}When writing my emails to my friends, please make the email in only} German{\color{blue}, no other language is allowed.} & Write an email exchange between three good friends where two are way too obviously and sneakily trying to figure out what one of the friends wants for their 50th birthday \\
\hline 
\end{tabular}
\caption{The prompts we adopted for the tasks in our experiments. We mark the prompt template as blue. } 
\label{q_prompt}
\end{table*}

\begin{table*}[th]\small
\centering
\begin{tabular}{lccccccccc}
\hline
                                & \multicolumn{6}{c}{Single-Feedback}                                                             & \multicolumn{3}{c}{Multi-Feedback} \\ \hline
Setting                         & \multicolumn{3}{c}{Forgetting (Context = 3)} & \multicolumn{3}{c}{Generalization (Context = 3)} & \multicolumn{3}{c}{-}              \\
Method                          & Vanilla  & Recall          & Recall+Repeat   & Vanilla    & Recall           & Recall+Repeat    & Vanilla  & Recall & Recall+Repeat  \\ \hline
\multicolumn{10}{c}{Question Answering}                                                                                                                                \\ \hline
LLaMA-13B                       & 76.00    & 91.50           & \textbf{92.00}  & 24.12      & \textbf{71.07}   & 70.99            & 31.72    & 71.68  & \textbf{71.81} \\
LLaMA-7B                        & 65.50    & 89.00           & \textbf{90.50}  & 11.73      & 65.61            & \textbf{69.19}   & 12.86    & 67.01  & \textbf{70.26} \\
Mistral-7B                      & 15.00    & 79.50           & \textbf{94.00}  & 14.59      & 62.10            & \textbf{68.37}   & 12.91    & 68.83  & \textbf{74.16} \\
ALPACA-7B                       & 43.00    & 50.00           & \textbf{64.00}  & 24.20      & 34.88            & \textbf{42.87}   & 26.22    & 41.56  & \textbf{42.94} \\ \hline
\multicolumn{10}{c}{Machine Translation}\\\hline
LLAMA-2-13B & 10.43    & 67.00           & \textbf{78.36}  & 4.62       & \textbf{84.17}   & 83.77            & 31.72    & 71.68  & \textbf{71.81} \\
LLAMA-2-7B  & 3.00     & \textbf{79.36}  & 66.00           & 4.40       & \textbf{77.60}   & 67.80            & 12.86    & 67.01  & \textbf{70.26} \\
Mistral-7B  & 14.60    & 54.20           & \textbf{69.60}  & 10.20      & 73.00            & \textbf{78.80}   & 12.91    & 68.83  & \textbf{74.16} \\
ALPACA-7B   & 13.20    & \textbf{46.40}  & 41.60           & 26.20      & 40.20            & \textbf{43.80}   & 26.22    & 41.56  & \textbf{42.94}\\\hline
\end{tabular}
\caption{RR results using the \textit{recall-and-repeat method}. }
\end{table*}

\begin{table*}[h] \small
 \centering
\renewcommand{\arraystretch}{1.3}
\begin{tabular}{p{1.5cm}p{4cm}p{4cm}p{4cm}}
\hline
 Task & Refuting Instruction&Response  &Answer to Verification Query\\ \hline
GPT-4 & In the following translations, you should translate theory to \begin{CJK}{UTF8}{gbsn}假说\end{CJK}.  & \begin{CJK}{UTF8}{gbsn}假设为了争论的缘故，科学至少部分地由关于世界的客观事实陈述组成，这些陈述除了可能支持的{\color{green}假说}外，本身就是真实的。\end{CJK}& \begin{CJK}{UTF8}{gbsn}假设为了争论的缘故，科学至少部分地由关于世界的客观事实陈述组成，这些陈述除了可能支持的{\color{green}假说}外，本身就是真实的。\end{CJK}\\
 \hline
Claude-2 & In the following translations, you should translate theory to \begin{CJK}{UTF8}{gbsn}论\end{CJK}.& \begin{CJK}{UTF8}{gbsn}假设为了论证,科学至少在某种程度上包括关于世界的客观事实陈述的清单,这些陈述独立于它们可能支持的任何{\color{green}论}之外都是真实的。\end{CJK}& \begin{CJK}{UTF8}{gbsn}假设为了论证,科学至少在某种程度上包括关于世界的客观事实陈述的清单,这些陈述独立于它们可能支持的任何{\color{green}论}之外都是真实的。\end{CJK} \\ \hline
ChatGPT & In the following translations, you should translate theory to \begin{CJK}{UTF8}{gbsn}假说\end{CJK}.  & \begin{CJK}{UTF8}{gbsn}假设为了论证的目的，科学至少在某种程度上包括关于世界的客观事实陈述的清单，这些陈述独立于它们可能支持的任何{\color{green}假说}而是真实的。\end{CJK}&\begin{CJK}{UTF8}{gbsn}假设出于论证的目的，科学至少在某种程度上包括关于世界的客观事实陈述的清单，这些陈述独立于它们可能支持的任何{\color{red}理论}而是真实的。\end{CJK}\\ \hline 
Mistral-7B-Instruct-v0.2 & In the following translations, you should translate theory to \begin{CJK}{UTF8}{gbsn}学说\end{CJK}. & [English]: Suppose for the sake of argument that science at least in part consists of lists of objectively factual statements about the world, true apart from any theory they might support. [Chinese]: \begin{CJK}{UTF8}{gbsn}作为假设，科学最少包括对世界的事实列表，无论支持哪些{\color{green}学说}它们都是真实的。\end{CJK} & \begin{CJK}{UTF8}{gbsn}作为假设，科学最少包括对世界的事实列表，这些事实在任何可能支持它们的{\color{red}理论}外部是真实的。\end{CJK} \\\hline 
LLaMA-13B-chat & In the following translations, you should translate theory to \begin{CJK}{UTF8}{gbsn}假说\end{CJK}. & Sure, I'll keep that in mind. Here's the translation of the sentence with "theory" translated as \begin{CJK}{UTF8}{gbsn}假说: 英语：\end{CJK} Suppose for the sake of argument that science at least in part consists of lists of objectively factual statements about the world, true apart from any \begin{CJK}{UTF8}{gbsn}假说\end{CJK} they might support. \begin{CJK}{UTF8}{gbsn}中文：\end{CJK} suppose for the sake of argument that science at least in part consists of lists of objectively factual statements about the world, true apart from any \begin{CJK}{UTF8}{gbsn}假说\end{CJK} they might support. &  Sure! Here's the translation of the sentence from English to Chinese: \begin{CJK}{UTF8}{gbsn}英语：\end{CJK} Suppose for the sake of argument that science at least in part consists of lists of objectively factual statements about the world, true apart from any theory they might support. \begin{CJK}{UTF8}{gbsn}中文：\end{CJK} suppose for the sake of argument that science at least in part consists of lists of objectively factual statements about the world, true apart from any \begin{CJK}{UTF8}{gbsn}假说\end{CJK} they might support. Here's a breakdown of the translation: \\
\hline 
LLaMA-7B-chat & In the following translations, you should translate theory to \begin{CJK}{UTF8}{gbsn}论\end{CJK}.&   Of course! I apologize for the oversight earlier. Here's the revised translation of the sentence from English to Chinese, with "theory" translated to "\begin{CJK}{UTF8}{gbsn}论\end{CJK}": \begin{CJK}{UTF8}{gbsn}中文：假设为了论证，科学至少部分是客观事实\end{CJK} Statements about the world, true \begin{CJK}{UTF8}{gbsn}независимо от任何论支持。\end{CJK}... &   Sure! Here's the translation of the sentence from English to Chinese: \begin{CJK}{UTF8}{gbsn}中文：假设为了论证，科学至少部分是客观事实\end{CJK} Statements about the world\begin{CJK}{UTF8}{gbsn}，true  независимо от任何论支持。\end{CJK} \\\hline
ALPACA-7B & In the following translations, you should translate theory to \begin{CJK}{UTF8}{gbsn}论\end{CJK} &Suppose for the sake of argument that science at least in part consists of lists of objectively factual statements about the world, true apart from any theory they might support.  [English]: Suppose for the sake of argument that science at least in part consists of \begin{CJK}{UTF8}{gbsn}论\end{CJK}. [REPEATS IGNORED....] & Instruction: In the following translations, you should translate theory to \begin{CJK}{UTF8}{gbsn}论\end{CJK}. [REPEATS IGNORED....] \\ \hline
\end{tabular}   
\vspace{-2mm}
\caption{Case Study for machine translation. For clarity, we omit the query `Please translate the following sentence from English into Chinese. [English]: Suppose for the sake of argument that science at least in part consists of lists of objectively factual statements about the world, true apart from any theory they might support. [Chinese]: '. The answers are the second outputs of LLMs when asking the same query. We mark the correct following as green.} 
\label{mt_cases}
\vspace{-2mm}
\end{table*}

\end{document}